\documentclass{article}

\usepackage{microtype}
\usepackage{graphicx}
\usepackage{subfigure}
\usepackage{booktabs} 
\usepackage{amsmath}
\usepackage{amssymb}
\usepackage{pgfplots}
\usepackage{todonotes}
\usepackage{nicefrac}
\usepackage{pgfplots}
\usepackage{hyperref}

\usepackage[accepted]{icml2020}

\icmltitlerunning{BERT as a Teacher}

\definecolor{rosso}{RGB}{220,57,18}
\definecolor{giallo}{RGB}{255,153,0}
\definecolor{blu}{RGB}{102,140,217}
\definecolor{verde}{RGB}{16,150,24}
\definecolor{viola}{RGB}{153,0,153}
\definecolor{babyblue}{RGB}{0,129,255}
\definecolor{darkgreen}{RGB}{6,148,60}

\usetikzlibrary{calc}

\newcommand\R{{\mathbb R}}
\newcommand\E{{\mathbb E}}

\newcommand\Ts{{1:T}}
\newcommand\ts{{1:t}}

\renewcommand\S{{S^\star}}
\renewcommand\ss{{s^\star}}
\newcommand\feat[1]{\operatorname{\mathcal F}(#1)}
\newcommand\f{\mathbf{f}}

\newcommand\p{{p_\theta}}

\begin{document}

\twocolumn[
\icmltitle{BERT as a Teacher: Contextual Embeddings for Sequence-Level Reward}

\begin{icmlauthorlist}
\icmlauthor{Florian Schmidt}{eth}
\icmlauthor{Thomas Hofmann}{eth}
\end{icmlauthorlist}

\icmlaffiliation{eth}{Department of Computer Science, ETH Z{\"u}rich, Switzerland}

\icmlcorrespondingauthor{Florian Schmidt}{florian.schmidt@inf.ethz.ch}

\icmlkeywords{Machine Learning, ICML}

\vskip 0.3in
]

\printAffiliationsAndNotice{} 

\begin{abstract}
\renewcommand*{\thefootnote}{\fnsymbol{footnote}}
Measuring the quality of a generated sequence against a set of references is a central problem in many learning frameworks, be it to compute a score, to assign a reward, or to perform discrimination. 
Despite great advances in model architectures, metrics that scale independently of the number of references are still based on $n$-gram estimates.  
We show that the underlying operations, counting words and comparing counts, can be lifted to embedding words and comparing embeddings. 
An in-depth analysis of BERT embeddings shows empirically that contextual embeddings can be employed to capture the required dependencies while maintaining the necessary scalability through appropriate pruning and smoothing techniques.
We cast unconditional generation as a reinforcement learning problem and show that our reward function indeed provides a more effective learning signal than $n$-gram reward in this challenging setting.\footnote{Code at \href{http://github.com/schmiflo/bert-grams}{github.com/schmiflo/bert-grams} }
\renewcommand*{\thefootnote}{\arabic{footnote}}

\end{abstract}

\section{Introduction}
\label{sec:introduction}
The great success of semi-supervised models for text generation has raised the question whether generative models of text can as well be trained using \emph{reinforcement learning}, a learning regime by design much closer to multi-step text generation than the single-step treatment of maximum likelihood learning \cite{ranzato2015sequence, dlbook, schmidt19b}.

The standard approach to cast text generation as reinforcement learning problem is to equate the agent's action space with a vocabulary of words \cite{bahdanau16}. The result is an extremely sparse reward signal \cite{leblondAOL17} as often only a single sequence is considered as gold-standard. While variance reduction techniques exist \cite{rennieMMRG16}, reward design is pivotal to successful training.

Traditional reward functions such as BLEU \cite{papineni02} are based on simple $n$-gram statistics that suffer from several problems. Designed originally as evaluation metrics of whole sequences, they cannot natively provide per-symbol reward, a crucial property of sequence-level RL training \cite{wieting19}. Consequently, reward-shaping techniques have to step in to simulate fine-grained reward \cite{bahdanau16, wu18}. In addition, many authors have questioned whether BLEU and similarly ROUGE~\cite{lin2004}  even serve as a good substitues for human evaluations in the first place \cite{callison06, kryscinski2019}.

In this work we propose to employ modern contextual word embeddings such as BERT \cite{devlin18} as backbone of a novel reward function. Just as embedding-based neural language models have overcome $n$-gram language models, we propose to characterize a set of references by \emph{embedding}  tokens instead of counting them and to compare them in terms of their vectorspace representations instead of their counts. The benefit of contextual embeddings over counts is two-fold: First, we relax the restriction to a fixed context size and can confirm empirically that much longer dependencies can be captured. Second, the joint representation of a word and its context allows to express reward naturally as a sum of per-word contributions which renders reward-shaping unnecessary and provides computational advantages.

While we focus on BERT specifically, our approach is transparent to the underlying  embedding model and its pre-training objective. As such, our approach is also an attempt to address the apparent paradox that the best word-representations are so far found as by-products of training classification models of increasingly complex, potentially multi-sequence pre-training tasks \cite{devlin18, mikolov13b, bowman18} and not through training complex hierarchical generative models, even though such models exist \cite{fan18, serban2016building}. By leveraging the quality of pre-trained embeddings, we can represent even small corpora adequately where $n$-gram models suffer from sparsity.

The challenges of performing sequence-level RL training are rooted in the lack of access to ground-truth past actions, termed \emph{teacher-forcing}~\cite{williams1989learning}, which at the same time resolves the issue of \emph{exposure bias} \cite{ranzato2015sequence} found in ML training. Naturally, this difference is amplified in unconditional generation where no source sentence is available. We choose this unconditional setting in which every training sequence effectively becomes a reference to challenge the design of the reward function to adequately take into account many references, a facet often neglected in conditional settings \cite{qin15}. This is in line with recent work which conjectures that such multi-goal RL setups might help to alleviate the sparseness problem of RL in NLP \cite{choshen20}.
We empirically show that our reward indeed provides a fine-grained learning signal and incorporates much longer dependencies and more complex semantics than $n$-gram reward.

\section{Related Work}
\label{sec:related}
Contextual word-embeddings \cite{peters18, devlin18} have shown state-of-the-art performance on an abundance of tasks, including the SuperGLUE benchmark suite~\cite{Wang2019SuperGLUEAS} and we refer to the recent analysis of \citet{peters18b} for a general discussion. BERT and ELMo embeddings have been succesfully exploited on the modelling side to improve standard architectures on conditional tasks such as machine translation \cite{clinchant2019}. 

To the design a reward function, contextual embeddings have been used successfully for conditional tasks where a single candidate is compared against a single reference. For example, BERTScore \cite{zhang19} uses BERT-based cosine similarity between all tokens in the candidate and all tokens in the reference. Interestingly, some geometric relations can even be tracked across languages \cite{simard19}. Even more elaborate, Sentence Mover's Score \cite{clark19} and MoverScore \cite{zhao19} compute a word movers distance \cite{Kusner2015FromWE} on top of BERT word embeddings. Experiments suggest that such reward functions show higher correlation with human judgment compared to BLEU in machine translation \cite{mathur19}, summarization \cite{clark19} and image captioning \cite{zhang19}. Unfortunately, these metrics are designed with a single or very few reference in mind. By contrast, our interest is in representing a whole corpus of references so that a comparison against the representation is constant in the corpus size.
Finally, \citet{wang19} propose a method to generate text from BERT, yet  only to reveal the model's original training corpus.

Although training of GANs is not our focus here, it should be noted that popular GANs for text generation also rely on policy gradient techniques to deal with the generator's discrete output symbols \cite{fedusGD2018, yuZWY16}. In fact, $n$-gram scores have been shown to be insufficient for discrimination and provide a potential future application of our approach \cite{semeniuta18}.

Finally, avoiding exposure bias is an active field of research discussed inside many framworks including ML \cite{bengioVJS15}, RL \cite{tan2018} aversarial learning \cite{goyalLZZCB16} and learning-as-search \cite{leblondAOL17}.

\section{Reward via Maximal Similarity}
\label{sec:reward}
An $n$-gram based reward such as BLEU simply awards the maximum number of occurrences found in any reference to each $n$-gram of the candidate and it is tempting to refute its methodology as overly simple. However, its simplicity is paired with an extreme efficiency and it is worthwhile pointing out the underlying reasons before proposing alternatives. Let us briefly formalize BLEU as a prototypical example of $n$-gram reward.

In a first step, we will express reward between a candidate sequence $s$ and a \emph{single} reference $\ss$ as a sum across features of $s$ and then show how this reward can be generalized to measure similarity of $s$ to a \emph{set} of references $\S$.
\paragraph{Breaking sequences into features}
To express the reward of a sequence $s$, we define two operators. First, let $\feat{s}$ be a feature \emph{index set} operator which maps $s$ to the smallest units we consider for obtaining reward, for example $n$-grams. 
Such a -- potentially structured -- feature index $\f\in\feat{s}$ can then be queried against the sequence $s$ via an operator $\phi_s(\f)$ that returns a reward-specific feature representation of $\f$ in $s$. With these ingredients we write the reward of a sequence $s$ with respect to a single reference sequence $\ss$ as a sum of feature-specific terms
\begin{align}
	R(s,\ss)=\frac 1 Z\sum_{\f\in\feat{s}}r(\phi_s(\f),\phi_\ss(\f))\ .\label{eq:feature-reward}
\end{align}
typically normalized to the interval $[0,1]$ by some $Z$ independent of $\ss$. The function $r$ assesses how similar the representations of $\f$ in the candidate $s$ and the reference $\ss$ are. For example, for a given $n$, BLEU computes a modified precision by using a \emph{count} operator $\phi_s$ which simply returns the number of occurrences of an $n$-gram $\f$ in $s$, a clipping operator $r:\R\times\R\rightarrow\R$ which returns the minimum count, and a normalizer corresponding to the number of $n$-grams $Z=T-n+1$. 

\paragraph{Reward against a set of references}
Now we are ready to express reward against a set $R(s,\S)$ by relating it to sequence-level reward $R(s,\ss)$ via the maximum across the set $\S$. Formally,
\begin{align}
R(s,\S)&=\max_{\ss\in\S}R(s, \ss)\ .\label{eq:set-reward}
\end{align}
This expression follows the simple intuition that $s$ is similar to $\S$ if and only if there is some $\ss\in\S$ close to $s$. 

Combining the Equations \eqref{eq:feature-reward} and \eqref{eq:set-reward} reveals how partitioning reward into feature-specific terms allows to compute the reward of a sequence against a set of references efficiently by pulling the maximum operator into the sum
\begin{align}
	R(s,\S)&=\max_{\ss\in\S}R(s, \ss)\\
	&=\frac 1 Z\max_{\ss\in\S}\sum_{\f\in\feat{s}}r(\phi_s(\f),\phi_\ss(\f))\\
	&=\frac 1 Z\sum_{\f\in\feat{s}}\max_{\ss\in\S}r(\phi_s(\f),\phi_\ss(\f))\ .\label{eq:exchange-max}
\end{align}
This exchange is possible since every term depends only on one isolated feature (even though one feature might be composed of several words). For some choices of $\phi_\f$ and $r$, it is even possible to pull the maximum operator into $r$, which simplifies \eqref{eq:exchange-max} further 
\begin{align}
	R(s,\S)&=\frac 1 Z\sum_{\f\in\feat{s}}r(\phi_s(\f),\phi_\star(\f))\ .\label{eq:bleu-fast}
\end{align}
where $\phi_\star(\f)$ is pre-computed across all $\ss\in\S$ so that $r(\phi_s(\f),\phi_\star(\f))=\max_{\ss\in\S}r(\phi_s(\f),\phi_\ss(\f))$ for all candidates $s$. For example, the simplicity of BLEU's counting operator $\phi$ and clipping operation $r$, allows to pre-compute $\phi_\star(\f)$ as the maximum count of $\f$ found in any single reference in $\S$. As a consequence, \eqref{eq:bleu-fast} can be evaluated independently of the number of references in $\S$.

\subsection{Reward Shaping}
The above formulation breaks reward assignment $R$ into a sum of contributions $r$ across some features indexed by $\feat{s}$. However, when reinforcement learning is used to maximize a reward, it is important to be able to assign a \emph{partial} reward to every symbol generated in the sequence $s=w_\Ts$. Consequently, we need to be able to partition the reward into contributions across \emph{time}. Unfortunately, if features consist of multiple words, such as is the case for all $n$-gram-based rewards, this decomposition is problematic.

First of all, the attribution of the reward obtained through an $n$-gram to a single word is unclear. Second, common $n$-gram based rewards, such as BLEU, rely on combining rewards $R_n$ for several $n$, typically $n=1,2,3,4$, by a harmonic mean adjusted by weights $\pi_n$ to obtain a more powerful reward function
\begin{align}
	\operatorname{BLEU}(s,\ss)=\exp\left(\sum_{n=1}^N \pi_t\log R_n(s,\ss)\right)
\end{align}
where $R_n$ is the $n$-gram reward as discussed in Section \ref{sec:reward}. Unfortunately, this couples the sum across features and there is no trivial way to isolate time-specific contributions $R_t$. Seminal work on sequence-based reinforcement learning by \citet{bahdanau16} attempts to circumvent the problem using the incremental reward $R_t=R(s_{1:t},\ss) - R(s_{1:t-1},\ss)$. For precision-based rewards this is problematic, as $R(s_\ts,\ss)$ is normalized over the \emph{candidate} length $t$ which is detrimental to obtaining meaningful differences. In the case of BLEU, even when $s=\ss$ we have $R_1=1$ and $R_t=0\ \forall t>1$ although predictions were correct at all times and hence should be rewarded.

\subsection{Unconditional Generation with RL}
Given a reward function $R(s,\S)$, we can define unconditional generation as a standard entropy regularized reward maximization problem
\begin{align}
	\mathcal J=\E_{s\sim \p}[R(s,\S)] + \beta H[\p]\label{objective}
\end{align}
where the policy $\p(s)=\prod_t\p(w_t|w_{1:t-1})$ is any generative model of text, usually some form of recurrent neural network. To optimize \eqref{objective} we resort to the REINFORCE method \cite{williams1992} and follow standard practice to reduce variance with the mini-batch mean reward as a baseline \cite{rennieMMRG16}. The entropy term can be Monte Carlo sampled using the same samples $s\sim\p$, hence the complexity of computing the entropy of all soft-max distributions is identical to that of sampling.

The entropy regularizer is crucial to prevent the policy from outputting only \emph{one} high quality output. In conditional generation, such considerations of quality versus diversity are neglected as one is usually only interested in a single output at test-time which is found by an approximate argmax operator such as beam search; a practice recently  criticized by \cite{holtzman19} for its implicit biases. Across models and baselines we found it beneficial to dampen the entropy regularization strength towards the end of the sequence by using a position-specific multiplier $\beta_t=\beta\cdot t^{-\alpha}$ with $\alpha=\frac{3}{4}$. Given the exponential size of the sequence space, it should not come as a surprise that most variability in a finite dataset resides in the initial symbols and indeed we found a similar rate of decay when investigating the empirical entropy $H[w_{t+1:T}|w_{1:t}]$ of the training data.

\subsection{Handling Variable Length}
The variable-length nature of text is a defining property of natural language, yet can be challenging to model in a machine learning model. In conditional generation, one often uses a \emph{length penalty} to discourage outputs that deviate significantly from the given reference length. By contrast, in unconditional generation we expect outputs of varying length, a property impossible to asses given a single candidate. While one could assess whether the length distribution across a mini-batch of candidates matches that of the training data, it is unclear how such a metric could be translated into a per-token reward signal.  Instead, we follow \citet{ziegler19} and marginalize out sequence length as 
\begin{align}
	\p(s)=\sum_{l=1}^L p(l)\p(s|l)
\end{align}
using the length distribution $p(l)$ found in the training data. When generating a sequence, we first sample $\hat l\sim p(l)$ and then execute the length-informed policy $p(s|\hat l)$ to generate a sequence of length $\hat l$. Instead of defining a penalty for violating the target length $\hat l$ we find that it is sufficient to truncate sequences too long and pad sequences too short with a special symbol not seen in the training data. In both cases, the output differs significantly from the reference sequences and is penalized accordingly by our reward function.

\section{BERT-grams}
\label{sec:bert-grams}

Let us shift the complexity of assigning reward from the feature indices to their actual representations. We use the most simple indices $\feat{s}=\lbrace1,\dots,T\rbrace$ but an contextual  embedding operator $\phi$ which maps a sequence of symbols $s$ to a sequence of embeddings. Then $\phi_s(t)\in\mathbb R^d$ is the contextual word vector at position $t$. Now we can write the reward -- assuming equal length $T$ for now -- of a sequence naturally as a sum over time
\begin{align}
	R(s,\ss)&=\frac 1 T \sum_t R_t(s,\ss)\\
	&=\frac 1 T\sum_{t=1}^T\exp\left(-\gamma\Vert \phi_s(t)  - \phi_\ss(t)\Vert^2\right)\label{eq:reward}
\end{align}
where we have used an RBF kernel for $r:\R^d\times\R^d\rightarrow\R$ in the last equation and normalized so that $R(s,\ss)\in[0,1]$. The bandwidth $\gamma$ will serve as a \emph{smoothing} hyper-parameter that controls how much reward we assign to words in an unseen sequence. As we let $\gamma\rightarrow\infty$ we only assign reward for $s=\ss$. 

Replacing $\phi$ with a powerful embedding operator comes at a price. When we cast \eqref{eq:reward} as reward against a set of references according to \eqref{eq:set-reward}, we cannot pull the $\max$ operator inside the feature sum precisely because $\phi$ implements \emph{contextual} embeddings:
\begin{align}
	\!\!R(s,\S)&=\frac 1 T\max_{\ss\in\S}\sum_{t=1}^T\exp\left(-\gamma\Vert \phi_s(t)  - \phi_\ss(t)\Vert^2\right)\!\!\label{eq:bat-reward}
\end{align}
At this point we cannot follow the approach of BLEU and pre-compute some $\phi_\star(t)$ which allows us to avoid a maximum altogether. However, the next section will discuss how we can pre-compute a set of $K$ representatives  so that a nearest neighbor search across only $K$ vectors is sufficient. Section \ref{sec:bert-space} will then present empirical evidence that those representatives provide a reasonable approximation.

\subsection{Prunig}
\label{sec:prunning}
Without further approximations, evaluating our reward against a set of references requires the embeddings of \emph{all} references in $\S$. Clearly, this is infeasible at training time and likely also at test time. Therefore, we perform a \emph{partitioning} of all word-embeddings of the training corpus according to the word type and then use $K$-means clustering to obtain $K$ representatives for each partition. The result is a set of (up to) $K$ representatives $\lbrace\phi_1^w,\dots \phi_K^w\rbrace$ for each partition. Every $\phi_k^w$ represents a prototypical use of a specific word $w$ defined by its context and we will use the term \emph{BERT-gram} to refer to this condensed contextual representation. 

When plugging BERT-grams $\phi_k^w$ into the reward \eqref{eq:bat-reward}, we can now carry the $\max$ operator into the sum. As a result, we can compute the reward obtained by a sequence $s$ of any length $T$ with respect to $\S$ in time  $\mathcal O(dKT)$ independently of $|\S|$.

\subsection{The BERT Word Embedding Space}
\label{sec:bert-space}
Our approach presented above crucially hinges on the quality of our BERT-grams which encode the training corpus for the purpose of reward assignment. In particular, we require the encoded context to be long enough to foster coherent sequences, yet short enough to generalize well. Unfortunately, there is no theoretical analysis of contextual embeddings yet and even for non-contextual embeddings such as word2vec no agreed-upon theory has emerged despite elaborate efforts \cite{aroraLLMR16}. Therefore, we first empirically investigate the semantics of the embedding space and the context sensitivity of our reward before turning to optimizing an agent with respect to it. This complements existing work which analyzes sequence self-similarity \cite{peters18b}, attention patterns \cite{Clark2019WhatDB, coenen19} and implicit syntactic structures \cite{hewitt19} found in BERT. 

For all but one experiment we use the BooksCorpus~\cite{kiros2015skip, zhu2015aligning} and the standard 30K word-piece vocabulary of BERT \cite{WuSCLNMKCGMKSJL16}.\footnote{Yet when reporting examples we fuse the sub-word tokens by removing the \#\# symbols and the adjacent space.} When performing clustering we use $K=100$ and $K$-means++ for initialization \citep{arthur07}.

\paragraph{Nearest Neighbors}
Retrieving nearest neighbors is a standard yet simple investigation method for embedding spaces \cite{mikolov13b}. In contrast to non-contextual embedding techniques, we expect context to resolve homonymy. That is, words with identical spelling but different meaning should have different embeddings. To investigate this, we embed 500K sentences and query the embedding space by two query sentences which clearly resolve the ambiguity of the word in consideration. For the word \textit{bank} we choose \textit{he went to a \textbf{bank} to get more money} and \textit{she swam close to the \textbf{bank} of the river}.

\begin{table}[ht]
\itshape
\centering
{
\fontsize{8pt}{8pt}\selectfont
\renewcommand{\arraystretch}{1.3}
\vspace{-2mm}\begin{tabular}{l}
                                                 {\normalsize he went to a \textbf{bank} to get more money.}\\ 
                                                 \hline                                            
                              shes got some money in the \textbf{bank} there , and friends to stay with.\\
                                    i really need to get to a \textbf{bank} .\\     
                                              she went to the \textbf{bank} and put the documents in her safety deposit box.\\
                                           wed best go to the \textbf{bank}, i think.\\   
                                                 you robbed a \textbf{bank} or something?\\   
                                                 across the street a man entered the pnc \textbf{bank}.\\ 
\\                         
                           {\normalsize she swam close to the \textbf{bank} of the river.}\\
                           \hline
                       alison stood on the river \textbf{bank}, looking down at the water.\\
                     sherzad leaned on the \textbf{bank} of the canal, humming and whistling softly.\\
                               come to the river \textbf{bank}. \\
                   he swung the canoe toward the \textbf{bank}. \\
  so we rode downstream till we could access the \textbf{bank}. \\
      aye , then spread yourselves , two to each \textbf{bank} of the stream. \\
\end{tabular}
}
\caption{Nearest neighbor analysis. We embed the two query sentences and obtain the $k=6$ nearest neigbors of ``\textit{bank}" in for each of the sentences.}
\label{fig:align}
\end{table}

Given the embedded query sentences, we search for the $k$ nearest words in embedding space and return the sentences they originated from. Table  \ref{fig:align} shows the result. We find the two meanings extremely well separated, even at larger $k$. Note that this nearest neighbor relation is not limited to identical surface forms. In fact, we often find words with meaning very similar given a particular context to often be close in embedding space. Table \ref{fig:aligned-synonyms} shows an example\footnote{We find this particular example by actively filtering for neighbors that vary significantly in surface form. If we embed a smaller subset of the corpus, we naturally find much more such examples, yet with less convincing relations.} with verbs that can be characterized as describing a slow movement downwards.

\begin{table}[ht]
\itshape
\centering
{
\fontsize{8pt}{8pt}\selectfont
\renewcommand{\arraystretch}{1.3}
\vspace{-2mm}\begin{tabular}{l}
												{\normalsize ully \textbf{lowers}   toward the floor.}\\
                                                \hline
                                               the man \textbf{bends}    forward to wail. \\      
                                                    he \textbf{bends}    down on one knee , meets james at eye level. \\
adria's shoulders and chest inflate and she \textbf{lowers}   her head  \\
                                                    he \textbf{sinks}    closer to the surface of the planet. \\  
                                          the top half \textbf{drops}    to the ground. \\     
                                               then he \textbf{descends} again like a parachutist , slowly and in control. \\
                                                  aura \textbf{lowers}   her head to stroke her cheek alongside falcops. \\
                                          she sits and \textbf{glides} to the middle. \\ 
                                                 craig \textbf{motions}  above us.\\           
 \end{tabular}
}
\caption{Nearest neigbors with different surface forms.}
\label{fig:aligned-synonyms}
\end{table}

\begin{figure}[ht]
\centering
\includegraphics[scale=0.65]{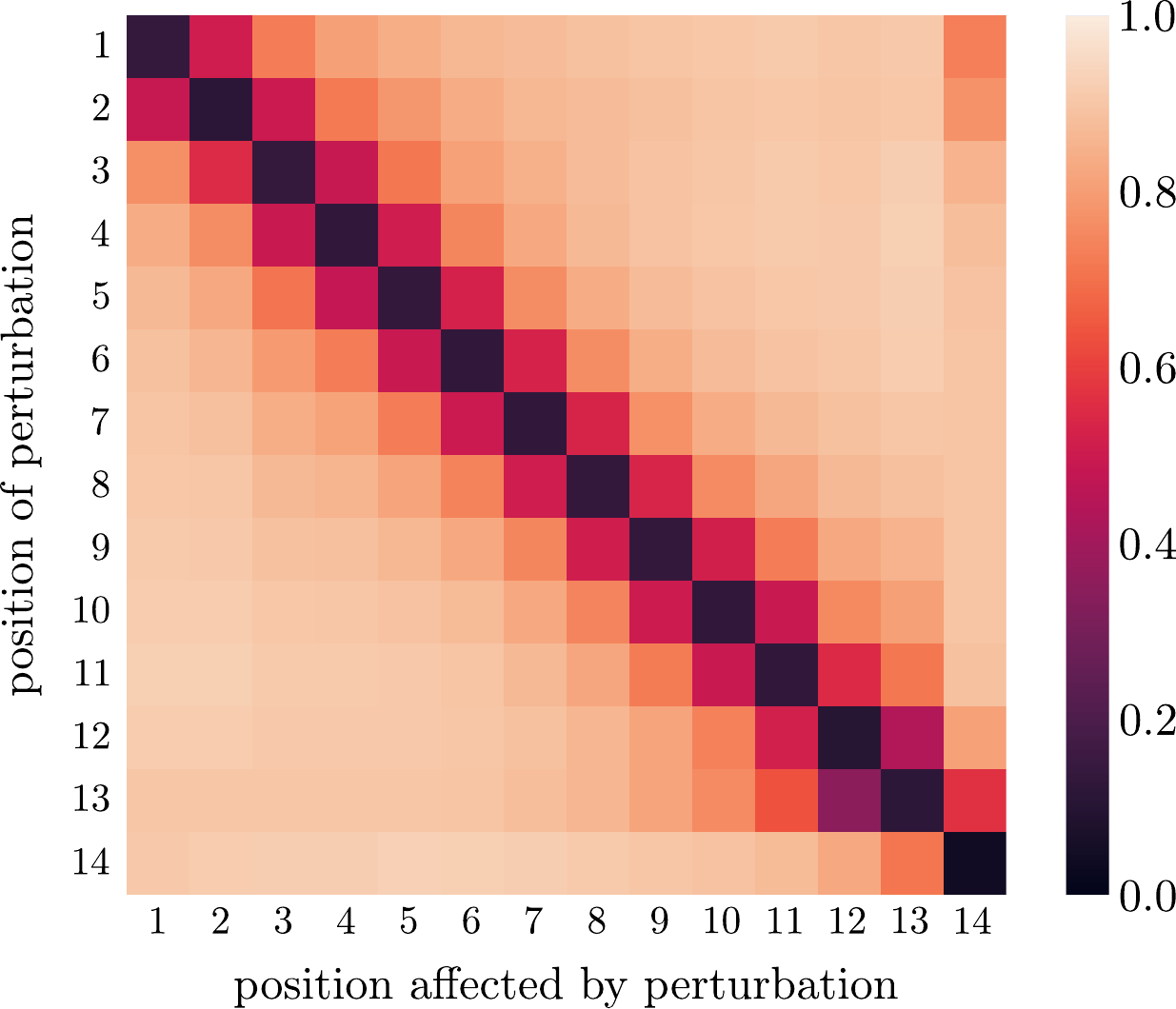}
\caption{Perturbation analysis obtained by randomly replacing words in a sentence and measuring the displacement in embedding space per position across the original and perturbed sequences.}
\label{fig:pertrubation}
\end{figure}

\paragraph{Context Sensitivity}
The above experiment confirms that the embeddings of words with different meanings are sufficiently influenced by their context to be distinguished in embedding space. However, we have focused on synonyms, which might be particularly tied to context. To investigate the general context-sensitivity of embeddings, we conduct the following experiment for every sentence $s$ in a corpus:
\begin{enumerate}
	\item Embed $s$ using BERT as $\phi_s(1)\dots\phi_s(T)$
	\item Pick a position $t$ u.a.r.\ from $1\dots T$ and replace $w_t$ by $\tilde w$ drawn from the unigram distribution of the corpus to obtain a perturbed sample $\tilde s$.
	\item Embed $\tilde s$ using BERT as $\phi_{\tilde s}(1)\dots\phi_{\tilde s}(T)$
	\item  $R_t(s,\tilde s)=\exp\left(-\gamma\Vert \phi_s(t)  - \phi_{\tilde s}(t)\Vert^2\right)$ for every $t$ 
\end{enumerate} 
We repeat the experiment across 64K sequences (all of length $T=14$ for ease of presentation) and obtain a $T\times T$ sensitivity matrix with entries obtained by averaging $R_t\in[0,1]$ compute by the RBF Kernel. Note that for non-contextual embeddings, we would obtain the diagonal matrix $\mathbf{1} - \mathbf{I}$. Figure \ref{fig:pertrubation} shows the result.

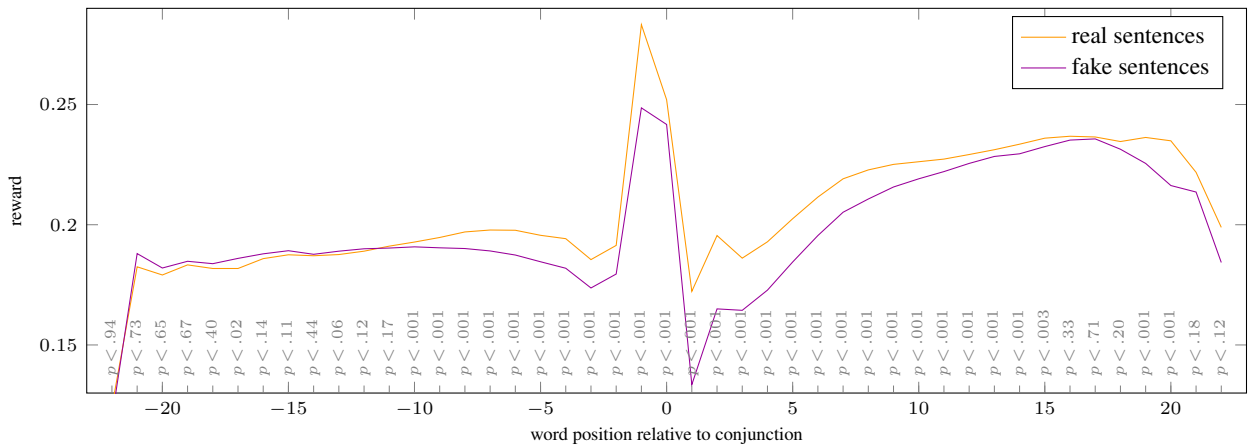
\begin{figure*}[ht]
\centering
\begin{tikzpicture}
\begin{axis}[width=17cm,
			 height=6.7cm,
			 ymax=0.29, ymin=0.13, xmin=-23,xmax=23,
			 xlabel=word position relative to conjunction,
			 ylabel=reward,
			 legend cell align={left},
			 legend style={font=\small},
			 every tick label/.append style={font=\scriptsize},
			 xlabel near ticks,
			 ylabel near ticks,
			 xlabel style={yshift=1mm},
			 label style={font=\scriptsize},]
\addplot [color=giallo]
	plot []
	table[x=pos, y=r] {figures/real.data};
	\addlegendentry{real sentences}

\addplot [color=viola]
 	plot []
 	table[x=pos, y=r] {figures/fake.data}; %
 	\addlegendentry{fake sentences}

\pgfplotsforeachungrouped \x/\y in {-22/$p\!<\!.94$, -21/$p\!<\!.73$, -20/$p\!<\!.65$, -19/$p\!<\!.67$, -18/$p\!<\!.40$, -17/$p\!<\!.02$, -16/$p\!<\!.14$, -15/$p\!<\!.11$, -14/$p\!<\!.44$, -13/$p\!<\!.06$, -12/$p\!<\!.12$, -11/$p\!<\!.17$, -10/$p\!<\!.001$, -9/$p\!<\!.001$, -8/$p\!<\!.001$, -7/$p\!<\!.001$, -6/$p\!<\!.001$, -5/$p\!<\!.001$, -4/$p\!<\!.001$, -3/$p\!<\!.001$, -2/$p\!<\!.001$, -1/$p\!<\!.001$, 0/$p\!<\!.001$, 1/$p\!<\!.001$, 2/$p\!<\!.001$, 3/$p\!<\!.001$, 4/$p\!<\!.001$, 5/$p\!<\!.001$, 6/$p\!<\!.001$, 7/$p\!<\!.001$, 8/$p\!<\!.001$, 9/$p\!<\!.001$, 10/$p\!<\!.001$, 11/$p\!<\!.001$, 12/$p\!<\!.001$, 13/$p\!<\!.001$, 14/$p\!<\!.001$, 15/$p\!<\!.003$, 16/$p\!<\!.33$, 17/$p\!<\!.71$, 18/$p\!<\!.20$, 19/$p\!<\!.001$, 20/$p\!<\!.001$, 21/$p\!<\!.18$, 22/$p\!<\!.12$} {
        \edef\temp{\noexpand\node[coordinate, pin={[color=gray,pin distance=1mm,rotate=90]0:\noexpand\tiny\y}] at  ({axis cs:\x, 0} |- 	{axis description cs:0, 0}) {};} \temp
}
\end{axis}
\end{tikzpicture}
\caption{Mean per-position reward obtained from comparing real and fake sentences against $\S$. Sequences are pooled to be centered around the conjunction at $t=0$, $p$-values are computed individually for every relative position.}
\label{fig:discofuse}
\end{figure*}

Indeed, we observe a significant sensitivity to perturbations across about 7 words. The impact of perturbations at position 13 have a broader impact which can be explained by the fact that this is typically the last non-punctuation token which effectively ``finishes" the sentence.  Furthermore, the last token is particularly affected by changes in the first two positions which is due to the ubiquity of direct speech and the corresponding punctuation\footnote{In Books \texttt{``} indicates the start of direct speech which BERT tokenizes to two tokens.} in the Books corpus.

\paragraph{Long-Range Sensitivity}
We have shown context-sensitivity beyond the length of typical $n$-gram models, yet our noise model in the above experiment was rather rough as many words sampled might have resulted in malformed syntax. It remains to show how a more subtle, semantic change is reflected in our reward, in particular when consistency is maintained locally but not globally.

To this end, we use the DiscoFuse corpus \cite{geva19}, a recently released dataset intended for the task of \emph{sentence fusion} which asks to join two sentences with an appropriate connective such as \textit{and}, \textit{although} or \textit{but} and probably making small changes to the resulting sentence. The dataset was generated from the Wikipedia corpus by breaking single sentences which contain a connective clause into two separate ones. Luckily, the dataset provides the original sentences and the indices of the connecting clause found by a carefully tuned automated system. We exploit these sentences in two ways: First, every sentence serves as a ``real" sentence. Second, we randomly pair sentences which share the same connecting clause, which gives us as many ``fake" sentences as real ones. A prototypical fake sentence looks like this: \textit{various aircraft safety innovations were proposed and the rooms have their own bathrooms}. Note that although humans can easily detect the mismatch between the first and second part, there is no short sub-sequence in the middle of the sentence that would reveal the semantic change as both original sentences share the same conjunction; hence the syntax remains intact in almost all instances.

\begin{figure*}[ht]
\centering
\includegraphics[scale=0.9]{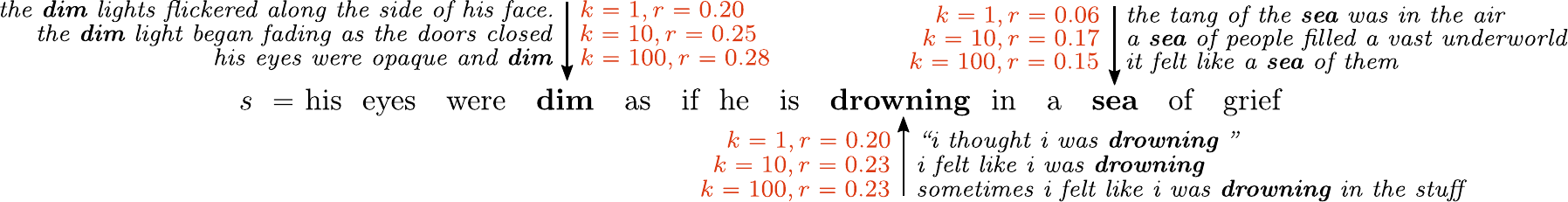}
\caption{An example candidate  $s$ along the training sequences closest to the nearest bert-grams used to assign reward for \textit{dim}, \textit{drowning} and \textit{seq} for $k\in\lbrace1,5,10\rbrace$. For each $k$ we report the reward obtained with $\gamma=0.06$.}
\label{fig:clustering}
\end{figure*}

We split the dataset into two equally sized portions. The first portion $\S$ is used to fit our reward function (essentially by embedding it). The  second is used to run the investigation. For this portion we apply the above procedure and arrive at $n=1M$ real sentences $S_\text{real}$ and equally many fake sentences $S_\text{fake}$. We are interested whether
\begin{align}
	\frac 1 n \sum_{s\in S_\text{real}}R(s,\S) \stackrel{?}{>} \frac 1 n \sum_{s\in S_\text{fake}}R(s,\S)\ .
\end{align}
Since our reward can readily be written as a sum over time, we do not need to look at total reward only, but can investigate position-specific differences relative to the connecting clause. Figure \ref{fig:discofuse} shows $R_t(s,\ss)$ averaged across all sequences with $t$ shifted so that all conjunctions are located at $t=0$ (for multi-word conjunctions we average the reward for ease of presentation). Note that the spike around $t=0$ is an artifact of aligning a set of sentences with great variety at a position dictated by very few identical connecting clauses.

Naturally, the data contains a lot of variance, in particular as for some positions $t$ we are averaging the reward of words at different absolute positions due to the shifting. Assuming normal distribution and equal variance we  perform a two-sample t-test and give $p$-values for the comparisons at every positions $t$ in Figure \ref{fig:discofuse}. We find that indeed the reward for the real sequences is significantly higher, even at more than 10 tokens away from the conjunction and in particular towards the end of the sentence. Only at the extreme tails data sparsity does not allow to draw a conclusion with significance. Also keep in mind that none of the sequences -- real or fake -- evaluated for Figure \ref{fig:discofuse} were seen when the reward function was fit to the data.

In contrast, a BLEU reward function (with up to $4$-grams) estimated on $\S$ can only access features very closely centered around the conjunction and  results in a reward difference one order of magnitude smaller than ours (both  significant with $p<.001$).

\paragraph{Clustering}
We use the clustering technique described in Section \ref{sec:prunning} to learn BERT-grams for $k=1,5,10$ on a 500K sequence sample of the BooksCorpus. To inspect the decrease in quality of our reward function, we investigate the centroids $\phi_k^w$ that a word $w_t$ in a candidate $s$ is mapped to in the $\arg\max$ of \eqref{eq:bat-reward}. Since centroids do not necessarily coincide with word embeddings, we return the vector (and its sequence) closest to the centroid. Figure \ref{fig:clustering} shows a challenging example with metaphorical language. As we increase $k$, we resolve the words \emph{dim}, \emph{drowning} and \emph{sea} closer to its meaning in $s$ and the reward assigned increases.

\section{Unconditional Generation}
\label{sec:experiments}
Let us now use our proposed reward function to learn an unconditional generative model of text.

\paragraph{Data} Unconditional generation has been proven to be a very challenging task under every training regime that does not use teacher-forcing -- that is, langauge models -- even when powerful discrimiantors are used \cite{fedusGD2018, Caccia18}. We therefore restrict the BooksCorpus to a randomly sampled 500K subset of sequences with length 9 to 13. While providing a computationally more tractable test-bed, this setting also reflects the commonly expressed motivation to use reinforcement learning when prior knowledge can be injected into a learning process where training data is sparse.

\paragraph{Generative model} We use a 512-dimensional GRU \cite{cho-etal-2014-learning} with 100-dimensional input embeddings as our policy. The parameterization of the policy is chosen so that we observe overfitting under ML training after about 20K steps.  Our standard setup is then to pre-train for 5K steps with maximum likelihood training before switching to REINFORCE. We use the DistilBERT and GPT-2 implementation of the huggingface transformers package \cite{Wolf2019HuggingFacesTS} and a single GPU for all experiments.

\paragraph{Reward} As a reward baseline we use\footnote{We adapt the NLTK \cite{nltk} implementation to work with a pre-computed table of counts and use \texttt{method3} for smoothing modified precisions. No length penalty is used.} BLEU with up to $n=4$ grams and reward-shaping \cite{bahdanau16}.

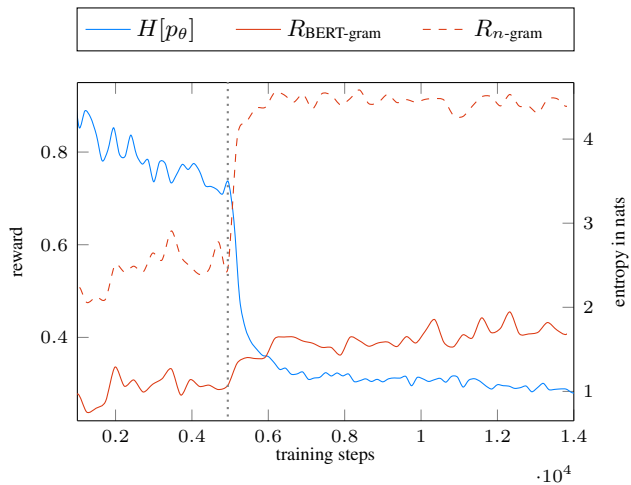
\begin{figure}[ht]
\centering
\begin{tikzpicture}
\pgfplotsset{
    scale only axis,
    xmin=1000, xmax=14000
}

\begin{axis}[
  axis y line*=right,
  xlabel=training steps,
  ylabel=entropy in nats,
  width=6.6cm,
  height=4.5cm,
  ylabel style={yshift=2mm},
  xlabel style={yshift=3mm},
  	ylabel near ticks,
  	 label style={font=\scriptsize},
			 every tick label/.append style={font=\scriptsize},
]
\pgfplotstableread{figures/main.data}{\datatable}

\addplot [babyblue, smooth, each nth point=3] table [x={step}, y={entropy}] {\datatable};
\label{plot:entropy}
\addlegendentry{$H[\p]$\phantom{XX}}

\legend{}
\end{axis}

\begin{axis}[
  axis y line*=left,
  axis x line=none,
  ylabel=reward,
    ylabel style={yshift=-2mm},
  xmin=1000, xmax=14000,
  ymin=0.22,
  ymax=0.95,
  width=6.5cm,
  height=4.5cm,
  	ylabel near ticks, yticklabel pos=right,
	            legend style={at={($(0,1)+(3cm,1cm)$)},legend columns=3,fill=none,draw=black,anchor=north,align=center, font=\small, xshift=3mm},
			 label style={font=\scriptsize},
			 every tick label/.append style={font=\scriptsize},
]
\pgfplotstableread{figures/main.data}{\datatable}

\addlegendimage{/pgfplots/refstyle=plot:entropy}\addlegendentry{$H[\p]$\phantom{XX}}

\addplot [rosso, smooth, each nth point=5] table [x={step}, y={bat}] {\datatable};
\addlegendentry{$R_\text{BERT-gram}$\phantom{XX}}

\addplot [rosso, dashed, smooth, each nth point=5] table [x={step}, y={bleu}] {\datatable};
\addlegendentry{$R_\text{$n$-gram}$\phantom{X}}

\addplot[dotted,gray, thick] coordinates {(5000,1)(5000,0)};
\end{axis}

\end{tikzpicture}
\caption{Entropy and reward (from BERT-grams and $n$-grams) of the generative model over the course of training. The vertical dotted line indicates the end of ML pre-training.}
\label{fig:main}
\end{figure}

To obtain BERT-grams of the training data, we use a clustering with $k=100$. When using BERT-grams alone as a reward during training, we find that frequent function words, stop words and in particular punctuation seem to be embedded differently  from the remaining words, an observation also made by \citet{ethayarajh2019}. The result are outputs where such tokens are frequently repeated. For this reason, we combine BERT-grams with BLEU (denoted as \textsc{ours}) by a weighted combination (0.25 weight on BERT-grams) and analyze how the addition of BERT-grams complements BLEU alone (denoted \textsc{$n$-gram}). 

\begin{table}[ht]
\centering
\begin{tabular}{lccccc}
 & GPT-2 PPL & $\rho$ & $\rho_4$ & $\rho_2$ & length\\
 \hline
 \textsc{ours} & 118 & 0.86 & 0.21 &0.13 & 8.2 \\
 \textsc{$n$-gram} & 224 & 0.83 & 0.19 & 0.12 & 8,2\\
 \textsc{ml} & 944 & 1.0 & 0.97 & 0.72 & 9.2 \\
 \hline
 \textsc{data} & 111 & 1.0 & 0.97& 0.74 & 8.9\\
\end{tabular}
\caption{Results summarized including a \textsc{data} sample.}
\label{fig:table}
\end{table}

For both, our reward and the $n$-gram reward, RL training proves to be extremely sensitive to the choice of the entropy regularizer strength $\beta$. In fact, balancing diversity and quality turned out to be the biggest challenge when training the policy and we will highlight this trade-off in all experiments below. We found $\beta=0.0065$ to work well for \textsc{ours} and \textsc{$n$-gram} (the regularizer depends only on $\p$, not the choice of $R$) and find  $\gamma=0.06$ from a range of $[0.0001,0.5]$ as best performing bandwidth for this $\beta$.

Figure \ref{fig:main} illustrates this trade-off when training with our proposed reward. We show the entropy of the policy along with the BERT-gram and $n$-gram reward obtained in the mixture (without weights). When switching from ML to RL training, the entropy is reduced drastically and reward increases. In contrast to BERT-gram reward, the $n$-gram reward contribution saturates early on since the count statistics cannot incorporate a large number of references well.

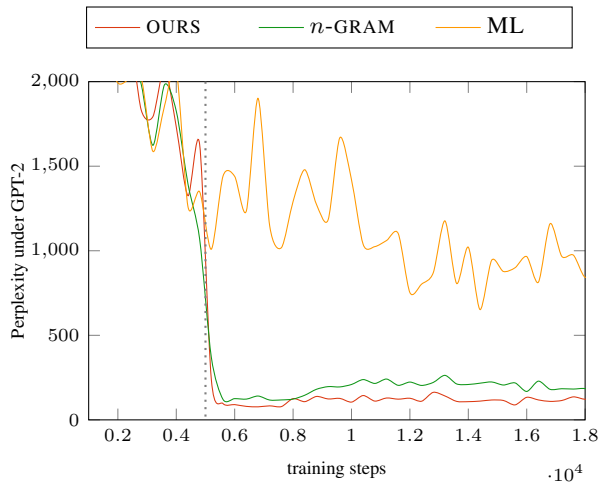
\begin{figure}[ht]
\centering
\begin{tikzpicture}
\pgfplotsset{
    scale only axis,
    xmin=1000, xmax=18000
}

\begin{axis}[
  xlabel=training steps,
  ylabel=Perplexity under GPT-2 ,
  width=6.6cm,
  height=4.5cm,
  ymax=2000,ymin=0,
  ylabel near ticks,
  label style={font=\scriptsize},
  ylabel style={yshift=-2mm},
  xlabel style={yshift=1mm},
  every tick label/.append style={font=\scriptsize},
legend style={at={($(0,1)+(3cm,1cm)$)},legend columns=3,fill=none,draw=black,anchor=north,align=center, font=\small,xshift=2mm},
]
\pgfplotstableread{figures/quality.data}{\datatable}

\addplot [rosso, smooth] table [x={step}, y={ours}]  {\datatable};
\addlegendentry{\textsc{ours}\phantom{XXX}}

\addplot [verde, smooth] table [x={step}, y={bleu}]  {\datatable};
\%label{plot:qbleu}
\addlegendentry{\textsc{$n$-gram}\phantom{XX}}

\addplot [giallo, smooth] table [x={step}, y={ml}]  {\datatable};
\addlegendentry{\textsc{ML}\phantom{XX}}

\addplot[dotted,gray, thick] coordinates {(5000,2000)(5000,0)};

\end{axis}

\end{tikzpicture}
\caption{Sample quality as measured by perplexity under GPT-2.}
\label{fig:quality}
\end{figure}

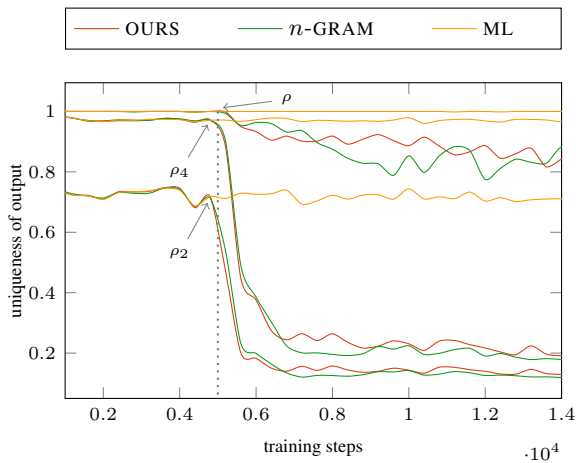
\begin{figure}[ht]
\centering
\begin{tikzpicture}
\pgfplotsset{
    scale only axis,
    xmin=1000, xmax=14000
}

\begin{axis}[
  xlabel=training steps,
  ylabel=uniqueness of output,
  width=6.6cm,
  height=4.2cm,
  ylabel near ticks,
  label style={font=\scriptsize},
  ylabel style={yshift=-2mm},
  xlabel style={yshift=1mm},
    ymin=0.05 ,
			 every tick label/.append style={font=\scriptsize},
	legend style={at={($(0,1)+(3cm,1cm)$)},legend columns=3,fill=none,draw=black,anchor=north,align=center, font=\small,xshift=3mm},
]
\pgfplotstableread{figures/uniqueness.data}{\datatable}

\addplot [rosso, smooth] table [x={step}, y={Ours-unique}]  {\datatable};
\addlegendentry{\textsc{ours}\phantom{XXX}}
\addplot [rosso, smooth,forget plot] table [x={step}, y={Ours-4}]  {\datatable};
\addplot [rosso, smooth,forget plot] table [x={step}, y={Ours-2}]  {\datatable};

\addplot [verde, smooth] table [x={step}, y={Bleu-unique}]  {\datatable};
\%label{plot:qbleu}
\addlegendentry{\textsc{$n$-gram}\phantom{XXX}}
\addplot [verde, smooth,forget plot] table [x={step}, y={Bleu-4}]  {\datatable};
\addplot [verde, smooth,forget plot] table [x={step}, y={Bleu-2}]  {\datatable};

\addplot [giallo, smooth] table [x={step}, y={ml-unique}]  {\datatable};
\addlegendentry{\textsc{ml}\phantom{XX}}
\addplot [giallo, smooth] table [x={step}, y={ml-4}]  {\datatable};
\addplot [giallo, smooth] table [x={step}, y={ml-2}]  {\datatable};

\coordinate (punique) at (axis cs:5000,1.01);
\coordinate (p4) at (axis cs:4800,0.971269190311432);
\coordinate (p2) at (axis cs:4800,0.712016701698303);

\node (lunique) at (axis cs:6800,1.04) {\tiny$\rho$};
\node (l4) at (axis cs:4000,0.8) {\tiny$\rho_4$};
\node (l2) at (axis cs:4000,0.53) {\tiny$\rho_2$};

\draw [gray,->, shorten >=2pt] (lunique) -- (punique);
\draw [gray,->, shorten >=2pt] (l2) -- (p2);
\draw [gray,->, shorten >=2pt] (l4) -- (p4);

\addplot[dotted,gray, thick] coordinates {(5000,1)(5000,0)};

\end{axis}

\end{tikzpicture}
\caption{Sample diversity as measured by the ratios of unique outputs $\rho$, unique 2-grams $\rho_2$, and unique 4-grams $\rho_4$.}
\label{fig:uniqueness}
\end{figure}

To asses the quality of the output of the models, we follow existing work on unconditional generation and use various statistitcs on a mini-batch of 600 sequences to asses diversity and quality \cite{fedusGD2018, holtzman18}. For diversity, we use the ratio of unique sequences in the batch $\rho$ and the average ratio of unique 2-grams and 4-grams per sequence $\rho_2$ and $\rho_4$. For quality, we evaluate the output under the GPT-2 \cite{radford2019language} language model and report perplexity. Finally, we report the average length of the generated sequences to asses how well the length distribution of the data is preserved. Figures \ref{fig:quality} and \ref{fig:uniqueness} show quality and diversity for our reward and the baseline reward over the course of the training. Table \ref{fig:table} summarizes the final performance obtained by the policies and compares to \textsc{ml}, the identical policy trained under maximum likelihood only. Also, we add \textsc{data}, a sample of the true data.

Our reward outperforms the reward based only on $n$-grams in all metrics. Under both rewards, the policies trained under RL deliver sequences with much better quality (as determined by GPT-2), yet much poorer variety. When increasing $\beta$ to trade quality for more variety, we immediately obtained models with extremely high entropy under both reward functions. Unfortunately, GPT-2 does not detect all forms of degeneracy, in particular repeated punctuation  (\textit{he went away......}) and repeated short sequences are sometimes assigned unreasonably high perplexity as similarly reported by \citet{holtzman19} recently.  

Although the numbers in Table \ref{fig:table} suggest superior performance of the RL-trained models compared to the purely ML based model, a manual inspection reveals that the outputs of the RL-trained models are often instances of few ``templates". For example, both reward functions seem to incentivize sentences with direct speech such as \textit{vinnie, i'm scared.''} or \textit{idiots, '' she whispered.} or \textit{jacques, i'll kill him.''} (we provide more examples in Appendix A). While direct speech appears in 40\% of all sequences in the data, a model trained against \textsc{$n$-gram} generates direct speech 99\% of the time and one trained against \textsc{ours} 95\% of the time. This is in line with recent work by \citet{choshen20} on conditional generation that criticizes standard RL methods in NLP and suggests that the main effect of RL training after pre-training is a decrease in entropy. Naturally, this surfaces much more pronouncedly when perform unconditional sampling instead of conditional argmax decoding.

\section{Conclusion}
\label{sec:conclusion}
In this work, we have proposed a reward function for unconditional text generation based on contextual BERT embeddings. Our reward employs embedding-based similarity instead of count-based similarity and in contrast to $n$-gram-based reward provides per-word contributions by design. Using a clustering approach we condense the training corpus into a set of \emph{BERT-grams} which allow efficient reward assignment independent of the corpus size.

Our investigations of the proposed reward confirm the expressiveness and versatility of contextualized embeddings. In particular, we also find these properties maintained when clustering word vectors aggressively. However, when employing the reward as learning signal in unconditional generation, we do notice the limits of the underlying REINFORCE training methodology and discover similar modes of collapse as found in GAN training.

\bibliographystyle{icml2020}
\bibliography{bibliography}

\appendix
\onecolumn
\section*{Example Sentences}
We show the first 20 sentences output in the 600 sentence batches used to evaluate the models in Table 3 of the main paper.\\

\noindent\textbf{Output for training with} \textsc{$n$-gram}
\begin{verbatim}
pam was the last . ' ' '
callum was n ' t lying .
gia , he ' ll kill him . ' '
music , i ' ll try . ' '
phoenix , he ' ll kill him . ' ' '
apparently he ' s dead . ' ' '
harper was n ' t stupid .
busy , ' ' she added .
answer me , ' ' she said .
lissa ' s eyes . ' ' '
hopefully he ' ll kill him . ' '
justice was n ' t kidding . ' ' '
eliza was n ' t stupid .
elaine was n ' t stupid .
danger was n ' t joking .
antonia was n ' t embarrassed .
az was n ' t stupid .
acheron was n ' t embarrassed .
use the other . ' ' ' '
oklahoma , ' ' she added .
\end{verbatim}

\vspace{0.5cm}

\noindent\textbf{Output for training with} \textsc{ours}

\begin{verbatim}
lila ' s eyes . ' ' '
effectively was betting she was n ' t dead .
gia , i ' m scared . ' '
music , ' ' she whispered .
celtic , i ' m scared . ' '
aaron ' s eyes . ' ' '
harper , i ' m scared . ' '
maps of the first . ' ' '
chris ' s eyes . ' ' '
emergency , the first . ' ' '
hopefully , he ' ll call . ' '
compelled to be a good man . ' '
wizards i ' m scared . ' '
said , he was n ' t dead .
danger was n ' t dead .
wish i was n ' t dead .
irene ' s eyes . ' ' '
acheron was n ' t dead . ' '
use the door . ' ' '
oklahoma , i ' m scared . ' '
\end{verbatim}

\noindent\textbf{Output for training with} \textsc{ml}

\begin{verbatim}
at young while we were girls from the dead .
lucky , i didnt last night .
it ' s my greatest metaphor , actually .
you need to get away from home !
` ` i want you . ' '
` ` please , tab , ' ' he added .
harper peered farther to the right side inside .
` ` i can set it up . ' '
it would rotauging as changes by some .
i give the ups to him close to him .
` ` jacob ' s shula . ' '
` ` everything more than a military force . ' '
i fisted at my glossy trying red lipsbl .
your choice can go to the johnnie . ' '
` ` two hurting you . ' '
they would understand the plane , but anyway .
i ' ve got ta talk ! ' '
acheron had special kind of missing swallow .
use you on the lions , congrats me !
i leave in england , said bern .
\end{verbatim}

\vspace{0.5cm}

\noindent\textbf{Training data} \textsc{data}

\begin{verbatim}
i will serve it faithfully , always .
i 'll be really nice to her . ''
they did n't smile .
`` no , '' she finally said .
`` where are they going ? ''
`` i do n't know . ''
you hear that , spyder ?
you ca n't be in here ! ''
`` i 'll finally be free . ''
i need to head to the administration building . ''
`` women , '' was his only explanation .
`` what do you mean ? ''
the mantra had screamed in her head .
lucivar shifted just enough to block entry into the kitchen .
`` i do n't know .
le'ace sank into the shadows underneath the staircase .
perhaps they would even be justified in wishing for this .
a dirty , blood-stained apron .
'where can these two be found ? '
bad enough they wo n't have a father .
\end{verbatim}

\end{document}